\documentclass[11pt]{article}

\usepackage[final]{acl}

\usepackage{times}
\usepackage{latexsym}

\usepackage[T1]{fontenc}

\usepackage[utf8]{inputenc}

\usepackage{microtype}

\usepackage{inconsolata}

\usepackage{graphicx}

\usepackage{amsmath}
\usepackage{array}
\usepackage{booktabs} 

\title{LeMAJ (Legal LLM-as-a-Judge): Bridging Legal Reasoning and LLM Evaluation}

\author{
    \textbf{Joseph Enguehard\textsuperscript{1}},
    \textbf{Morgane Van Ermengem\textsuperscript{1}},
    \textbf{Kate Atkinson\textsuperscript{1}},
    \textbf{Sujeong Cha\textsuperscript{2}},
    \\
    \textbf{Arijit Ghosh Chowdhury\textsuperscript{2}},
    \textbf{Prashanth Kallur Ramaswamy\textsuperscript{2}},
    \textbf{Jeremy Roghair\textsuperscript{2}},
    \textbf{Hannah R Marlowe\textsuperscript{2}},
    \\
    \textbf{Carina Suzana Negreanu\textsuperscript{1}},
    \textbf{Kitty Boxall\textsuperscript{1}},
    \textbf{Diana Mincu\textsuperscript{1}}
\\
    \textsuperscript{1}Robin AI,
    \textsuperscript{2}Amazon Web Services,
}

\begin{document}
\maketitle
\maketitle

\begin{abstract}
Evaluating large language model (LLM) outputs in the legal domain presents unique challenges due to the complex and nuanced nature of legal analysis. Current evaluation approaches either depend on reference data, which is costly to produce, or use standardized assessment methods, both of which have significant limitations for legal applications.

Although LLM-as-a-Judge has emerged as a promising evaluation technique, its reliability and effectiveness in legal contexts depend heavily on evaluation processes unique to the legal industry and how trustworthy the evaluation appears to the human legal expert. This is where existing evaluation methods currently fail and exhibit considerable variability.

This paper aims to close the gap: a) we break down lengthy responses into "Legal Data Points" (LDPs) — self-contained units of information — and introduce a novel, reference-free evaluation methodology that reflects how lawyers evaluate legal answers; b) we demonstrate that our method outperforms a variety of baselines on both our proprietary dataset and an open-source dataset (LegalBench); c) we show how our method correlates more closely with human expert evaluations and helps improve inter-annotator agreement; and finally d) we open source our Legal Data Points for a subset of LegalBench used in our experiments, allowing the research community to replicate our results and advance research in this vital area of LLM evaluation on legal question-answering.

\end{abstract}

\section{Introduction}

\begin{figure*}[t]
    \centering
    \includegraphics[width=1\linewidth]{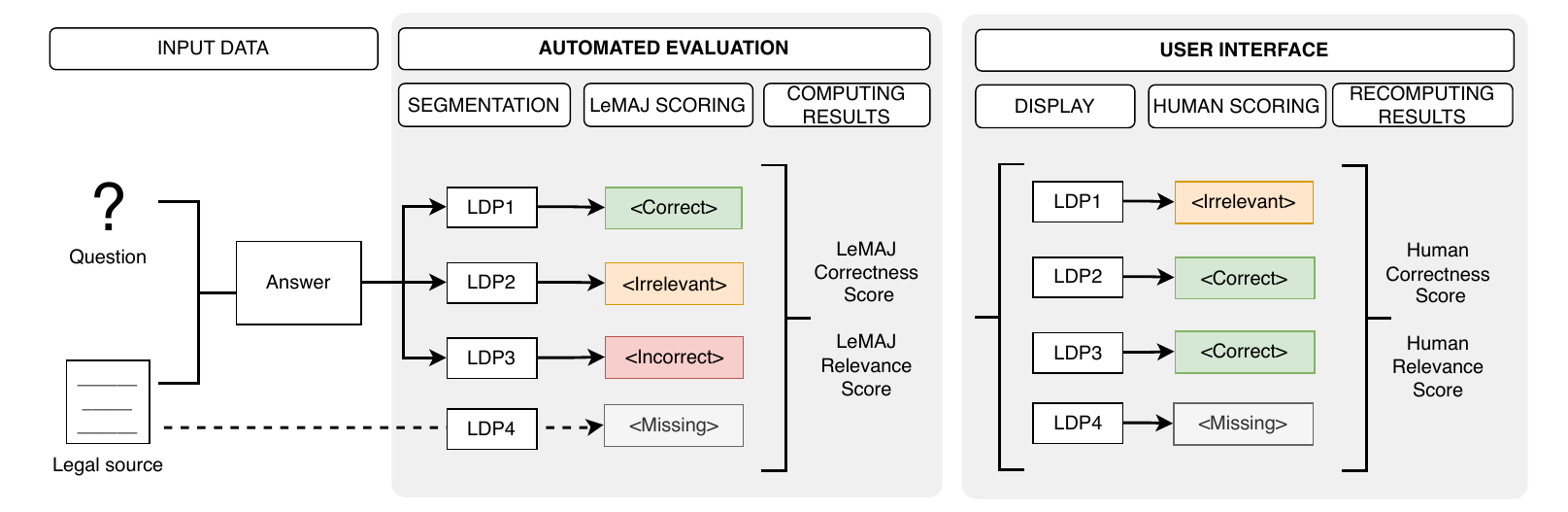}
    \caption{Based on a legal document, a question and an answer, our LeMAJ framework performs an automated evaluation by segmenting the answer into Legal Data Points (LDPs) and evaluating each one. A domain expert might also use this framework to manually evaluate each LDP and produce their own scores.}
    \label{fig:flow-diagram}
\end{figure*}

Large language models (LLMs) are increasingly integrated into legal workflows, supporting tasks such as contract analysis \citep{narendra2024enhancing}, document classification \citep{prasad2024exploring}, information extraction \citep{bommarito2021lexnlp} and court case outcome prediction \citep{chalkidis2019neural}.  Lawyers and legal professionals are relying more and more on legal outputs generated by an LLM to counter increasing pressure to 'do more for less' and this shows no sign of slowing down as the quality of legal AI tools increases \citep{gartner-ai}. 

Among these applications, question-answering over legal documents has emerged as a particularly valuable task. Legal evaluation (commonly referred to as 'legal review' in legal circles) is a crucial component of legal tasks, as it ensures that legal outputs are correct, complete and relevant. However, there are many challenges plaguing the evaluation of such outputs; among other things, it is time-consuming, requires highly specialized, expensive human experts and is prone to subjectivity  \citep{sun2024lalaevalholistichumanevaluation, gu2025surveyllmasajudge, rastogi2024insightsdisagreementpatternsmultimodal, guha2023legalbench}.

Traditional automated evaluation methods, which rely on high-quality reference answers ('ground truths') \citep{lin2004rouge, zhang2019bertscore}, have partially mitigated these challenges. Yet, their reliance on multiple expert lawyers to construct comprehensive ground truths \citep{wang2023maud} significantly limits their scalability, particularly in the legal domain, where the nuance and variability of legal language amplifies the requirements for large, representative datasets for reliable real-world evaluation.

LLM-as-a-Judge has emerged as a promising evaluation paradigm that can assess outputs without reference data and provide explanatory feedback \citep{zheng2023judging, fu2023gptscore}. However, these techniques can rely on hard to verify assessment approaches and can be prone to bias and task variability \citep{bavaresco2024llmsinsteadhumanjudges}. Existing question-answering methods that assess the whole answer as a unit~\citep{ryu2023retrieval}  fail to capture the granular assessment process that legal professionals employ \citep{pagnoni-etal-2021-understanding}. Recent research~\citep{krumdick2025no} has also shown that LLM-as-a-Judge methods perform significantly worse without good quality references. 

Our research reveals a critical insight: automated evaluations correlate significantly better with human expert judgments, even without human references, when they mirror how lawyers actually evaluate answers. From interviews with legal experts, we understand that there is added complexity and logic to legal reasoning that needs to be built into the LLM-as-a-Judge paradigm, bridging the gap between both methods for better alignment.

We propose LeMAJ (Legal LLM-as-a-Judge), a novel evaluation methodology specifically designed to emulate lawyers' evaluation processes. Our approach decomposes LLM-generated answers into discrete "Legal Data Points"—self-contained units of information—and systematically evaluates each for correctness and relevance while identifying critical omissions. This granular assessment provides the detailed feedback legal practitioners require beyond simple accuracy scores, drawing inspiration both from techniques in summary evaluation \citep{liu-etal-2023-revisiting, liu-etal-2023-towards-interpretable, tan-etal-2024-towards-automated} and from our own user study.
    
Through empirical studies, we first demonstrate that using LDPs improves alignment between human and LLM evaluation and correlation with gold-standard meta-reviews. We will present how our LeMAJ framework consists of two core elements: the LeMAJ automated evaluation based on LDP segmentation resulting in Correctness and Relevance scores, as well as a user interface, displaying LDPs for annotation by human legal experts. We then show how our LeMAJ automated evaluation substantially outperforms various LLM and non-LLM baselines when evaluated on both LegalBench~\citep{guha2023legalbench}—an open-source legal dataset—and proprietary legal data. Next, we show how the LeMAJ user interface component improved inter-annotator agreement between human legal experts when reviewing with LeMAJ. Finally, we introduce a commercial use case for the use of LeMAJ by showing time savings through the triaging of answers for review.

To sum up, the main contributions of this article are the following:
\begin{itemize}
    \item We introduce LeMAJ, a novel automated evaluation framework for question-answering on legal documents that mimics lawyers' reasoning process, without requiring reference data.
    \item We demonstrate that its performance is superior on both open-source and proprietary datasets when compared to other methods, improving alignment with human evaluation and inter-annotator agreement. 
    \item We provide a breakdown of time savings in a commercial use case.
    \item We open source our Legal Data Points for a subset of LegalBench used in our experiments, allowing the research community to replicate our results and advance research in this vital area of LLM evaluation.
\end{itemize}
\section{Related Work}

\textbf{Automatic evaluation:} There are a number of evaluation methods used in the NLP space - N-gram based methods such as ROUGE \citep{lin2004rouge} and BLEU \citep{bleu}, or token level model-based methods such as BertScore \citep{zhang2020bertscoreevaluatingtextgeneration} or BartScore~\citep{yuan2021bartscore}. However, these methods require a human reference.
We also look at LLM-as-a-Judge methods, such as DeepEval \citep{deepeval}, but while they can be used without references, they tend to perform poorly when used in a reference-free setting, especially when the LLM itself is not able to produce a correct answer~\citep{krumdick2025no}. Furthermore, model-based methods like LLM-as-a-Judge suffer from task variability \citep{li2024llmsasjudgescomprehensivesurveyllmbased, gu2025surveyllmasajudge}, meaning that one solution to fit all tasks is unlikely to exist and people often tend to customize any existing offering. This calls for adapting existing LLM-as-a-Judge methods to legal specific tasks such as legal Q\&A.

\textbf{Replicating the reasoning process:} Automated evaluation methods are usually evaluated by computing a correlation against human scores \citep{li2024llmsasjudgescomprehensivesurveyllmbased}. Given the reliance on correlation with humans, it becomes imperative that higher levels of inter-annotator agreement are reached. There are a few factors that can contribute to rating disagreements such as problems with the study setup, insufficient information provided on the rating scale, complex or highly subjective tasks, or even the annotator's background \citep{rastogi2024insightsdisagreementpatternsmultimodal}. As a result, being able to reduce these factors has become a big research area, especially in situations where the answers being evaluated are verbose. Work in this space has proposed dissecting the answer into easier to evaluate units \citep{liu-etal-2023-revisiting, tan-etal-2024-towards-automated}, splitting the evaluation pipeline into interpretable measures \citep{liu-etal-2023-towards-interpretable}, or in cases where a high-quality reference output is available, measuring the similarity between the two \citep{zha-etal-2023-alignscore}. But issues still persist as the actual measure is dependent on the task at hand, with some favoring factuality and reducing hallucinations \citep{min-etal-2023-factscore}, while others focus on correctness and human alignment. Therefore we identify a need for a flexible and easily adaptable metric.
\section{Method}
\label{section-method}

\subsection{Aligning with Human Legal Evaluation: Legal Data Points (LDPs)}

Our evaluation methodology, LeMAJ, draws inspiration from two sources: 1) the systematic evaluation process employed by legal professionals when reviewing legal answers, and 2) recent advances in automated summary evaluation techniques. We propose that legal answers, similar to summaries, consist of discrete informational units. This granular assessment approach aligns with recent innovations in summary evaluation methodologies \citep{liu-etal-2023-revisiting, liu-etal-2023-towards-interpretable, tan-etal-2024-towards-automated, min-etal-2023-factscore}, which similarly break down content into atomic units of information for comprehensive evaluation.

A key difference with summarization evaluation methods such as ACU~\citep{liu-etal-2023-towards-interpretable} however is that applying summarization techniques to legal Q\&A requires adding a step, since summarization and legal Q\&A outputs serve a different purpose. Q\&A requires all atomic units of information to create a whole and answer a specific question, whereas summarization units are a condensed list of information.

To better understand the legal reasoning process, we performed a user study with four lawyers, asking them two simple questions: how would they approach a legal review step-by-step, and what do they look out for in an answer to grade the answer as satisfactory? We present their answers in Appendix \ref{appendix-user-study}. Our findings show that legal professionals typically evaluate responses in a similar fashion:
\begin{itemize}
    \item Assessing answers by mentally decomposing them into individual assertions, sentences or by referencing the citation linked to each assertion. This would usually happen in an \textit{ad hoc} way, with lawyers either annotating directly on a document or splitting the individual assertions mentally as they progress through the answer.
    \item Verifying the factual correctness of each assertion (against a reference if available, e.g. a contract or a summary).
    \item Assessing the relevance of each factually correct assertion to the question at hand.
    \item Identifying any critical omissions from the source document, if available. 
\end{itemize}
For examples of LDPs, we refer to Appendix \ref{appendix-exemples-of-legal-data-point}.  We present a detailed breakdown of the evaluation process in the following section.

\subsection{Evaluation process} 

\begin{table*}[t]
    \small
    \centering  
    \begin{tabular}{|p{3cm}|p{5cm}|p{5cm}|}
        \hline
        \textbf{Question} & \textbf{Output} & \textbf{Evaluation} \\
        \hline
        What are the payment terms? & 
        "The payment terms depend on the schedule set forth in the applicable Statement of Services. If customer chooses to bill through an invoice, payment is due within 45 days after the customer receives the invoice, unless otherwise specified on the applicable Order Form (6.3). "& 
        {\textless \textbf{Correct}\textgreater Payment terms depend on the schedule set forth in the applicable Statement of Services.} 

        {\textless \textbf{Correct}\textgreater For invoices, payment must be received within 45 days after customer receives the invoice.} 
        
        {\textless \textbf{Irrelevant}\textgreater Different payment terms may be specified on the applicable Order Form.} 

        {\textless \textbf{Missing}\textgreater Services may be suspended in the event of payment delinquency, unless payment is suspended due to a dispute between the Parties.}\\
    \hline
    \end{tabular}
    \caption{\textbf{Payment Terms Evaluation}. The answer is split into LDPs, each individually assessed according to our tagging system.}
    \label{table:payment-terms}
\end{table*} 

Our LeMAJ methodology emulates the legal reasoning process described above through two stages. First, given the inputs of a legal document, a question, an answer and (optionally) a ground truth answer, we employ an LLM to decompose the answer into distinct assertions made within the answer, which we define as Legal Data Points. Then, within the same prompt, we ask the LLM to tag each of the LDPs using the following classification system: 
\begin{itemize}
    \item \textbf{Correctness}: first, if the LDP contains a factual error or a hallucination, it is marked as \textless Incorrect\textgreater; 
    \item \textbf{Relevance}: factually correct LDPs are then assessed on their relevance to the question asked. This can be a rather subjective assessment in the absence of good reference data. Through prompting techniques we are able to tweak our LLM to be more or less stringent in its interpretation of relevance, similarly to how different lawyers might assess this criterion. If the LDP is considered irrelevant to the question, it is marked \textless irrelevant\textgreater; in a standard confusion matrix, this would be similar to a false positive classification;
    \item \textbf{Correct and Relevant}: LDPs that are both factually accurate and relevant are marked as \textless correct\textgreater; similar to the true positive classification;
    \item \textbf{Critical Omissions}: missing information that should have been included in the answer is added as new LDPs and marked as \textless missing\textgreater; this is based on the false negative classification.
\end{itemize}

Table~\ref{table:payment-terms} illustrates this tagging process with a practical example, pulled from the payment terms in a Master Service Agreement. 

Based on the Legal Data Point classifications, we derive the following quantitative metrics:
\begin{itemize}
    \item $\text{Correctness} = \frac{\text{\# Correct LDPs}}{\text{\# Correct LDPs} + \text{\# Incorrect LDPs}}$, measures factual accuracy by penalizing errors and hallucinations
    \item $\text{Precision} = \frac{\text{\# Correct LDPs}}{\text{\# Correct LDPs} + \text{\# Irrelevant LDPs}}$, measures relevance by penalizing irrelevant content
    \item $\text{Recall} = \frac{\text{\# Correct LDPs}}{\text{\# Correct LDPs} + \text{\# Missing LDPs}}$, measures completeness by penalizing omissions
    \item $\text{F1} = \frac{2 \times \text{Precision} \times \text{Recall}}{\text{Precision } + \text{ Recall}}$, the classic F1 metric that balances precision and recall to provide an overall Relevance score
\end{itemize}

Importantly, this proposed metric is highly adaptable, enabling the scores to be adjusted according to the specific priorities of the legal evaluation task. For instance, when identifying missing information is critical, greater emphasis can be placed on Recall or Critical Omissions in the final score, offering a more contextually relevant and precise evaluation. 
\section{Experiments and Results}

\subsection{Datasets}

\textbf{Proprietary:} To evaluate our method's real-world effectiveness and refine our judge creation approach, we sourced an internal real-world dataset. It contains 9 distinct contract types, with a total of 5 different contracts per type, totaling 1000 pairs of Q\&As. Subject matter experts developed specific questions for each contract type, along with corresponding ground truth answers to enable accurate assessment and quick iterations. The breakdown for the total amount of questions split by training and test sets can be seen in Appendix~\ref{appendix-datasets}.

\textbf{LegalBench:} For the purpose of validating our method, we ran it on an open-source dataset, curated by a collaborative team across educational institutions under the auspices of Stanford University as a benchmark to test LLMs' legal reasoning in the English language \citep{guha2023legalbenchcollaborativelybuiltbenchmark}. We selected a subset of twelve contracts from this database at random and 150 questions that contained both relatively simple topics (e.g. governing law) and more complex ones (e.g. competitive restriction exceptions). A breakdown of the data set can be seen in Appendix~\ref{appendix-datasets}. We selected a subset only through our in-house legal experts, prioritizing more complex questions that would be a challenge to both evaluation methods and human review. Moreover, due to the low amount of LLM-generated incorrect answers, we augmented this dataset with 20 manually created incorrect or partially incorrect answers, bringing the size of this dataset to 170.

\subsection{Improving alignment with legal experts}

Our core claim is that using the LeMAJ automated evaluation improves alignment with human evaluation on legal question-answering. We test our method against several baselines on both our proprietary dataset and the LegalBench dataset by comparing the scores produced by each method to the scores produced by human legal experts. Each answer is evaluated based on two criteria, as defined in the Section \ref{section-method} above:
    \begin{itemize}
        \item \textbf{Correctness}: is the answer factually correct given the question?
        \item \textbf{Relevance (i.e F1)}: is the answer relevant to the question? Are there any critical omissions?
    \end{itemize}
Both scores are graded on scale between 0 and 1. In the absence of an industry standard scoring mechanism \citep{belz-etal-2023-non}, we computed a human score whereby the human legal expert attributes a score between 1 and 5 for relevance and correctness respectively of an answer which is then converted to 0, 0.25, 0.5, 0.75 and 1 (see Appendix~\ref{appendix-iaa-grading}). 

\subsubsection{Baselines}

We compare our method against the following non-LLM baselines which require a reference (ground truth):

\begin{itemize}
    \item \textbf{BLEU}~\citep{bleu} and \textbf{ROUGE}~\citep{lin2004rouge}: BLEU and ROUGE are evaluation metrics for generated text that analyze n-gram overlap between generated and reference texts. While BLEU focuses on precision with a brevity penalty, ROUGE emphasizes recall, with variants including ROUGE-1 (unigrams), ROUGE-2 (bigrams), and ROUGE-L (longest common subsequence).
    \item \textbf{BERTScore}~\citep{zhang2019bertscore} and \textbf{BARTScore}~\citep{yuan2021bartscore}: Advanced text evaluation metrics that leverage pre-trained language models to assess text generation quality. BERTScore computes token-level similarities using contextual BERT embeddings and cosine similarity, while BARTScore evaluates text by measuring the conditional log-likelihood between generated and reference texts using a BART sequence-to-sequence model.
\end{itemize}

On the other hand, we also compare our method against some out-of-the-box LLM-as-a-Judge methods~\citep{deepeval}. We use these methods reference-free, to draw a fair comparison against LeMAJ.

For Relevance, we use the following:
\begin{itemize}
    \item \textbf{Answer Relevancy}: This metric assesses how relevant an output is in relation to an input, specifically aiming to compare the relevancy of each part of the answer to the question. 
    \item \textbf{Faithfulness}: Faithfulness aims at assessing how much an answer factually aligns with a context (in this context the contract used to produce an answer).
\end{itemize}

For Correctness, we use:
\begin{itemize}
    \item \textbf{Correctness}: Correctness aims to assess how correct an answer is given a question, an answer and a contract (the context). To compute this score, we use the G-Eval method of DeepEval with a prompt we reproduce in Appendix \ref{appendix-deepeval-prompt}.
    \item \textbf{Hallucination}: This metric is an out-of-the-box method from DeepEval which aims at assessing whether an answer is factually correct given a contract.
\end{itemize}

\subsubsection{Metrics}

\begin{table*}[t]
\centering
\small
\begin{tabular}{|p{1.8cm}|p{3cm}p{1.8cm}|p{3cm}p{1.8cm}|}
\hline
& \multicolumn{2}{|c|}{\textbf{Proprietary Dataset}} & \multicolumn{2}{|c|}{\textbf{LegalBench}} \\
\textbf{Method} & \textbf{Pearson (p-value)} & \textbf{Bucketed Accuracy} & \textbf{Pearson (p-value)} & \textbf{Bucketed Accuracy} \\
\hline
BLEU-1 & 0.049 ($1.52 \times 10^{-1}$) & 0.05 & 0.142 ($6.48 \times 10^{-2}$) & 0.08 \\
BLEU-2 & 0.095 ($5.26 \times 10^{-3}$) & 0.04 & 0.207 ($6.75 \times 10^{-3}$) & 0.08 \\
BLEU-3 & 0.113 ($8.67 \times 10^{-4}$) & 0.03 & 0.241 ($1.55 \times 10^{-3}$) & 0.09 \\
BLEU-4 & 0.105 ($1.98 \times 10^{-3}$) & 0.03 & 0.248 ($1.09 \times 10^{-3}$) & 0.08 \\
ROUGE-1 & 0.095 ($5.11 \times 10^{-3}$) & 0.04 & 0.210 ($6.05 \times 10^{-3}$) & 0.09 \\
ROUGE-2 & 0.133 ($8.65 \times 10^{-5}$) & 0.03 & 0.229 ($2.70 \times 10^{-3}$) & 0.09 \\
ROUGE-L & 0.107 ($1.58 \times 10^{-3}$) & 0.04 & 0.193 ($1.16 \times 10^{-2}$) & 0.09 \\
BERTScore & 0.174 ($2.52 \times 10^{-7}$) & 0.02 & 0.055 ($4.78 \times 10^{-1}$) & 0.05 \\
BARTScore & 0.105 ($2.02 \times 10^{-3}$) & 0.02 & 0.205 ($7.40 \times 10^{-3}$) & 0.08 \\
\hline
DeepEval-Answer Relevancy & 0.000 ($9.92 \times 10^{-1}$) & 0.37 & 0.079 ($3.07 \times 10^{-1}$) & \textbf{0.45} \\
DeepEval-Faithfulness & 0.053 ($1.20 \times 10^{-1}$) & 0.48 & -0.130 ($9.22 \times 10^{-2}$) & 0.41 \\
\hline
LeMAJ & \textbf{0.370} ($1.46 \times 10^{-29}$) & \textbf{0.50} & \textbf{0.354} ($2.13 \times 10^{-6}$) & 0.35 \\
\hline
\end{tabular}
\caption{
    \textbf{Results on Proprietary Dataset and LegalBench, measuring relevancy.}
    The first set of methods (BLEU, ROUGE, BERT and BARTscore) use references, while the second set (DeepEval and LeMAJ) evaluate answers without human references.
    LeMAJ shows high performance on both datasets.
}
\label{tab:result-relevance}
\end{table*}

We evaluated our method and each baseline using two metrics:
\begin{itemize}
    \item \textbf{Pearson Correlation}: given a score produced by a baseline or LeMAJ, we compute its correlation with human gold standard scores.
    \item \textbf{Bucketed Accuracy}: We round each score to the nearest lower quarter point (0, 0.25, 0.5, 0.75 or 1) bringing the continuous scores to the same scale as the human review for accuracy correlation computation. We opted for this metric due to an issue with Pearson correlation \citep{elangovan2025correlationimpacthumanuncertainty}, where if many answers are fully correct or relevant, a single change in a prediction can swing the correlation very differently. This is particularly true for Correctness, as our answers are typically fully correct more than 90\% of the time, particularly on LegalBench.
\end{itemize}

  We introduce a third metric for our method only: \textbf{LeMAJ Alignment}. This metric is measured by: 
  (1) mapping each LDP annotated by LeMAJ to an LDP annotated by a human reviewer. This mapping is done using an OpenAI Embedding model; and
  (2) comparing the tags in each LDP pair to produce an accuracy score, where any remaining unmapped LDP is considered irrelevant if added by the LLM or missing if added by a human reviewer.
Due to the nature of this metric, it can only be computed on our method and is not used for comparisons.

\subsubsection{Results}

\textbf{Experiment set-up:} To ensure a fair comparison between LLM-based methods, we used the same LLM: Claude 3.5 Sonnet v2. Moreover, following~\citep{ye2024justice}, we used a different LLM (Claude 3.5 Sonnet v1) to generate the answers in order to avoid introducing "self-enhancement biases".

\textbf{Relevance:} We present our results when measuring Relevance in Table~\ref{tab:result-relevance}.
Overall, LeMAJ significantly outperforms both non-LLM and LLM methods on our proprietary dataset, despite requiring no references (compared with non-LLM methods).
LeMAJ also outperforms all other methods in terms of correlation with human scores on the LegalBench dataset.
While DeepEval methods achieve a higher Bucketed Accuracy, these methods actually do not provide accurate information: they tend to give a nearly perfect score to each answer, and since 48.2\% of the answers are fully relevant, DeepEval methods are correct around half of the time. This is confirmed by the low correlation (0.079) with human scores.

\textbf{Correctness:} We present our results on Correctness in Table~\ref{tab:results-correctness}.
LeMAJ outperforms all other methods on both our proprietary dataset and LegalBench, achieving a higher correlation with human evaluations than the baselines.

\begin{table*}[t]
\centering
\small
\begin{tabular}{|p{1.8cm}|p{3cm}p{1.8cm}|p{3cm}p{1.8cm}|}
\hline
\textbf{Method} & \multicolumn{2}{|c|}{Proprietary Dataset} & \multicolumn{2}{|c|}{LegalBench} \\
 & \textbf{Pearson (p-value)} & \textbf{Bucketed Accuracy} & \textbf{Pearson (p-value)} & \textbf{Bucketed Accuracy} \\
\hline
BLEU-1 & 0.104 ($1.52 \times 10^{-3}$) & 0.02 & 0.135 ($7.99 \times 10^{-2}$) & 0.05 \\
BLEU-2 & 0.116 ($6.15 \times 10^{-4}$) & 0.02 & 0.172 ($2.51 \times 10^{-2}$) & 0.06 \\
BLEU-3 & 0.111 ($9.92 \times 10^{-4}$) & 0.02 & 0.195 ($1.06 \times 10^{-2}$) & 0.08 \\
BLEU-4 & 0.090 ($7.78 \times 10^{-3}$) & 0.02 & 0.158 ($3.94 \times 10^{-2}$) & 0.08 \\
ROUGE-1 & 0.139 ($3.94 \times 10^{-5}$) & 0.01 & 0.167 ($2.97 \times 10^{-2}$) & 0.05 \\
ROUGE-2 & 0.128 ($1.46 \times 10^{-4}$) & 0.02 & 0.203 ($7.89 \times 10^{-3}$) & 0.07 \\
ROUGE-L & 0.131 ($1.05 \times 10^{-4}$) & 0.01 & 0.170 ($2.69 \times 10^{-2}$) & 0.06 \\
BERTScore & 0.164 ($1.11 \times 10^{-6}$) & 0.01 & 0.128 ($9.72 \times 10^{-2}$) & 0.02 \\
BARTScore & 0.074 ($2.91 \times 10^{-2}$) & 0.02 & 0.201 ($8.57 \times 10^{-3}$) & 0.08 \\
\hline
DeepEval-Correctness & 0.077 ($2.35 \times 10^{-2}$) & 0.43 & 0.018 ($8.13 \times 10^{-1}$) & 0.24 \\
DeepEval-Hallucination & 0.080 ($1.79 \times 10^{-2}$) & 0.04 & -0.001 ($9.95 \times 10^{-1}$) & 0.14 \\
\hline
LeMAJ & \textbf{0.259} ($7.54 \times 10^{-15}$) & \textbf{0.95} & \textbf{0.700} ($2.52 \times 10^{-26}$) & \textbf{0.88} \\
\hline
\end{tabular}
\caption{
    \textbf{Results on Proprietary Dataset and LegalBench, measuring correctness.}
    LeMAJ outperforms both DeepEval and non-LLM evaluation method on our proprietary dataset and on LegalBench.
}
\label{tab:results-correctness}
\end{table*}

\subsection{Reducing inter-annotator disagreement}

Through our research, we understood that in traditional evaluation methods there is a high degree of variability and subjectivity between different human reviewers \citep{rastogi2024insightsdisagreementpatternsmultimodal}. This is a barrier to the reproducibility of evaluation results \citep{belz-etal-2023-non}. Through the experiment below, we show that we can use the LeMAJ user interface to guide human legal experts by pre-determining the segmentation of information in order to reduce the risk of arbitrary misalignment between humans. To do so, we computed two different human scores: the first relies on the LeMAJ framework to break down an answer into LDPs, after which the human legal expert uses the user interface to assess each LDP and annotate it according to our tagging system outlined above (as represented in Figure \ref{fig:flow-diagram}). The second human score is manual and consists of a more rudimentary evaluation schema (in the absence of an industry standard scoring mechanism \citep{belz-etal-2023-non}), asking the human legal expert to assess the relevance and correctness of an answer on a 5-point scale (which is then converted to 0, 0.25, 0.5, 0.75 and 1, see Appendix~\ref{appendix-iaa-grading}). 

\textbf{Experiment setup:} We tested whether the segmentation into LDPs can improve IAA between human reviewers. We performed this experiment on the LegalBench open-source dataset only and compared the following evaluations:
\begin{itemize}
    \item the evaluation by two human legal experts using a 5-point scale, as outlined in Appendix~\ref{appendix-iaa-grading}; and
    \item the evaluation performed by human legal experts using LeMAJ, whereby human legal experts score every LDP using our mechanism outlined above.
\end{itemize}
Next, we measure the difference between both human legal experts' respective evaluations in their manual evaluations and in their evaluations using the LeMAJ tool.

\textbf{Results:} The average inter-annotator agreement between different reviewers improves by 11\% when evaluating outputs for Correctness, indicating that the reviewers are more aligned in their assessment when using LeMAJ (Table \ref{tab:contract_iaa_merged} below). This can be explained by the fact that Correctness evaluates the factuality of the LDPs, which is arguably less prone to subjectivity and legal interpretation.  
\begin{table*}[h]
    \centering
    \small
    \begin{tabular}{p{4.5cm}|c|p{1.2cm}p{1.2cm}|p{1.2cm}p{1.2cm}}
        \toprule
        \textbf{Contract Type} & \textbf{\# QA pairs} & \multicolumn{2}{c}{\textbf{Correctness}} & \multicolumn{2}{c}{\textbf{Relevance}} \\
        & & \textbf{IAA: manual} & \textbf{IAA: LeMAJ} & \textbf{IAA: manual} & \textbf{IAA: LeMAJ} \\
        \midrule
        Co-Promotion Agreement & 19 & 0.579 & 0.789 & 0.421 & 0.526 \\
        Consulting Agreement & 12 & 0.833 & 1 & 0.667 & 0.5 \\
        Cooperation Agreement & 10 & 0.7 & 0.7 & 0.4 & 0.5 \\
        Distributor Agreement & 17 & 1 & 0.824 & 0.588 & 0.765 \\
        Endorsement Agreement & 16 & 0.75 & 0.875 & 0.562 & 0.438 \\
        Intellectual Property Agreement & 8 & 1 & 1 & 0.5 & 0.625 \\
        License Agreement & 10 & 0.6 & 0.7 & 0.4 & 0.6 \\
        Licensing and Distribution Agreement & 5 & 1 & 1 & 0.8 & 0.2 \\
        Outsourcing Agreement & 15 & 0.867 & 1 & 0.467 & 0.267 \\
        Promotion Agreement & 17 & 0.765 & 1 & 0.471 & 0.471 \\
        Strategic Alliance Agreement & 11 & 0.636 & 0.909 & 0.546 & 0.818 \\
        Website Hosting Agreement & 10 & 0.6 & 0.8 & 0.8 & 0.7 \\
        \midrule
        \textbf{Total} & 150 & 0.77 & 0.88 & 0.53 & 0.54 \\
        \bottomrule
    \end{tabular}
    \caption{Inter-Annotator Agreement (IAA) on Correctness and Relevance}
    \label{tab:contract_iaa_merged}
\end{table*}

A different picture emerges when assessing Relevance; due to the inherently subjective nature of Relevance, reviewers still present low inter-annotator agreement to a similar degree as between the manual evaluations. We believe that this is to be expected, as this is a notoriously subjective assessment in the legal sphere and typically defined by a given task, as opposed to a broad industry standard. The added value of our approach in using LeMAJ is that we obtain a more granular picture of the elements that were considered irrelevant, making the assessment by the reviewer more transparent and auditable (addressing some of the concerns raised in \citep{pagnoni-etal-2021-understanding}), and generating actionable insights that can be used to tweak the prompting strategy further.

\subsection{Scaling evaluations}

Given the high computational cost of developing, deploying and maintaining large models as judges \citep{li2024llmsasjudgescomprehensivesurveyllmbased, gu2025surveyllmasajudge}, as well as the reduction in speed during inference, we explored various options to reduce the size of the model while maintaining performance.

We explored a) prompt optimization techniques, b) data augmentation, and c) an LLM Jury framework involving multiple fine-tuned models. We found that the LLM Jury framework was most performant, but when balancing against cost, fine-tuning with augmentation brought the best tradeoffs. More results can be found in Appendix \ref{appendix:lemaj_iterations}.
\section{Commercial use case: reduce human review efforts through triage}

Considering the significant amount of time, effort and resources required for human evaluations, if we can use LeMAJ to triage less controversial answers, then we can reserve review by human legal experts for the contentious or at-risk answers only. As our LeMAJ Alignment score, and in parallel confidence in the LeMAJ evaluation, increases, we can rely more and more on LeMAJ to detect those answers that do not need meticulous human expert review. 

We measure this by showing the potential time savings created through this triaging system. First, we track the time spent by human legal experts doing both the manual evaluations and the evaluations using the LeMAJ tool. Next, we use the results of the automated LeMAJ evaluation on both our proprietary dataset and the LegalBench dataset and apply a set of thresholds to triage all results that LeMAJ has given a Correctness score of 1 \textit{\textbf{and}} a Relevance score of at least 0.80 (for our proprietary dataset) and 0.85 (for the LegalBench dataset). This enables us to create a split between answers that should be flagged for review and answers that are cleared. Our findings show that this results in time savings of up to 50\% on our proprietary dataset and up to 30\% on LegalBench (Tables \ref{table:qa-human-time} and \ref{tab:evaluation_metrics} in Appendix \ref{appendix-commercia-application}). When running against production-level tools it enables organizations to bucket information for training and iteration, gives users confidence in what to review, and other use cases.

\section{Conclusion}

We introduced LeMAJ, a novel evaluation framework that seeks to closely emulate the legal reasoning process. We have shown that by splitting up a legal answer into single units of information (Legal Data Points) we can use LLMs to evaluate the Correctness and Relevance of a given answer at more granular level. The results of this methodology show a stronger correlation between the LeMAJ evaluation and a human gold standard than existing benchmarks on both our proprietary dataset and a subset of an open-source dataset, LegalBench. Additionally, we demonstrate that LeMAJ can improve inter-annotator agreement on Correctness, addressing a critical issue in the development of high-quality reference data and for the effective evaluation of assessment methods. Finally, we showcase the time savings in a practical application and a potential deployment pathway for LeMAJ.

Future work will look at increasing the accuracy of the method, improving its ability to detect incorrect and missing information even further, as well as extending the scalability work to a multi-agent framework that can detect the needs of a task and adapt the metric on the fly, attempting to improve issues around task variability in a single LLM-as-a-Judge framework.

\newpage
\bibliography{colm2025_conference}

\appendix
\begin{table*}[t]
\centering
\begin{tabular}{lccc}
\hline
\textbf{Contract Type} & \textbf{Training} & \textbf{Testing} & \textbf{Total} \\
\hline
Lease Agreements & 60 & 40 & 100 \\
Supplier Agreements & 60 & 40 & 100 \\
SaaS Agreements & 60 & 40 & 100 \\
Master Service Agreements (MSAs) & 60 & 40 & 100 \\
Limited Partnership Agreements (LPAs) & 60 & 60 & 120 \\
Side Letters & 66 & 88 & 154 \\
Shareholder's Agreements (SHAs) & 60 & 40 & 100 \\
Non-Disclosure Agreements (NDAs) & 60 & 40 & 100 \\
Sale and Purchase Agreements (SPAs) & 51 & 34 & 85 \\
\hline
\textbf{Total} & 537 & 422 & 959 \\
\hline
\end{tabular}
\caption{Distribution of Legal Contract Types in Proprietary Dataset}
\label{table:proprietary-stats}
\end{table*}

\begin{table*}[t]
\centering
\begin{tabular}{lc}
\hline
\textbf{Contract Type} & \textbf{Total} \\
\hline
Hosting Agreement & 10 \\
Cooperation Agreement & 10 \\
Promotion Agreements & 36 \\
Endorsement Agreement & 16 \\
Licensing, Distribution and Marketing Agreements & 32 \\
Outsourcing Agreement & 15 \\
Intellectual Property Agreement & 8 \\
Consulting Agreement & 12 \\
Strategic Alliance Agreement & 11 \\
\hline
\textbf{Total} & 150 \\
\hline
\end{tabular}
\caption{Distribution of Legal Contract Types in LegalBench Subset}
\label{table:legalbench-stats}
\end{table*}

\section{Appendix A. User study on lawyers' approach to evaluation}
\label{appendix-user-study}
We interviewed four lawyers of varying level of seniority (junior to 5 years Post-Qualification Experience or PQE) who are currently working in the legal industry and we asked them to answer the following two questions:
\begin{itemize}
    \item When you are reviewing the output of another lawyer, how would you describe your process step-by-step?
    \item What elements would you assess to check that an output is satisfactory?
\end{itemize}
Below is a breakdown of their answers.

\textbf{Lawyer 1: }
"I would read the answer and read the corresponding part of the contract. I would then compare each sentence of the answer to the relevant section of the contract and decide how correct that part of the answer is. I would check sentence-by-sentence that the answer is correct."

\textbf{Lawyer 2:}
“I would identify key elements that I would check, for example if there are 5 exceptions I check that they are all present in the output.”

\textbf{Lawyer 3: }
"Legal Analysis Check + Quality Control. - Verify accuracy, breadth of response and citations. How I review for accuracy is as follows:
\begin{enumerate}
  \item Read the report question and the answer provided.
  \item Is the answer directly answering the question?
  \item If so, is the content factually correct - click and read through the citation(s) where the information was pulled from
  \item If so, confirm that the report has not missed information about the topic from anywhere else in the contract. Do a Ctrl F search in the contract for key words relating to the question/answer.
  \item If the report has missed information from the contract, I would re-review the prompt and potentially amend it to be more prescriptive. If the prompt appears fine, I would manually add in the information from the contract into the report aligning to the answer type - if a summary, I would potentially use Claude to summarise concisely.\footnote{Author's note: the reviewer is referring to answers generated by an LLM}
  \item Once complete for every section, do a final 2 eye scroll of the contract confirming all key aspects have been included."
\end{enumerate}

\section{Appendix B. Inter-annotator agreement grading mechanism}
\label{appendix-iaa-grading}

\textbf{Lawyer 4:} 
"\textbf{Process}
\begin{itemize}
    \item \textbf{Substantive Legal Analysis} 
    \begin{itemize}
        \item Verify that the responses provided accurately reflect the underlying documents - clicking into each citation to check the response
        \item Identify any gaps in analysis that should be addressed, ensuring that the document has been read and references as a whole, taking into account the interrelation between separate provisions / definitions
        \item Ensure that the form of output provided aligns with the client’s needs
    \end{itemize}
    \item \textbf{Accuracy \& Consistency} 
    \begin{itemize}
        \item Confirm factual accuracy, ensuring no misstatements or oversights
        \item Ensure that the format of the output for each issue is consistent across each document
    \end{itemize}
    \item \textbf{Clarity \& Readability} 
    \begin{itemize}
        \item Evaluate if the language is clear, concise, and appropriate for the audience.
        \item Identify and eliminate unnecessary jargon or overly complex phrasing.
        \item Ensure any explanations are easy to follow for the intended reader.
    \end{itemize}
    \item \textbf{Formatting \& Citation Check} 
    \begin{itemize}
        \item Ensure consistency of formatting throughout
        \item Verify correct citation style
        \item Ensure the document meets firm or client formatting standards
    \end{itemize}
    \item \textbf{Final Review} 
    \begin{itemize}
        \item Proofread for grammar, spelling, and punctuation
        \item Cross-check all numerical or financial figures if applicable
        \item Ensure overall consistency and completeness
    \end{itemize}
\end{itemize}
\textbf{Assessment of Output:}
\begin{itemize}
    \item \textbf{Legal Accuracy} – All outputs accurately reflect the contents of each agreement
    \item \textbf{Citations} - all outputs contain accurate citations
    \item \textbf{Comprehensiveness} – All issues are addressed and the outputs take into account the document(s) as a whole
    \item \textbf{Clarity} – The language is clear, concise, and free of ambiguity.
    \item \textbf{Professionalism} – Proper and consistent formatting and citations."
\end{itemize}

\begin{table*}[htbp]
\centering
\begin{tabular}{|p{8cm}|p{3cm}|p{2.8cm}|}
\hline
\textbf{Error Category} & \textbf{Occurrences (count of total)} & \textbf{Occurrences (sum of total)} \\
\hline
Data points from answer missing from LeMAJ evaluation. These are errors whereby the evaluation by LeMAJ has missed data points from the answer into its evaluation. & 34 & 41 \\
\hline
The errors are due to LeMAJ splitting up a data point into further data points, causing the evaluation to no longer align with the reference data. & 34 & 34 \\
\hline
LeMAJ tagged the data point wrong. & 38 & 44 \\
\hline
LeMAJ is too lenient concerning level of detail when grading a data point. & 57 & 90 \\
\hline
Additional data points are added by LeMAJ that are not in the ground truth or answer. & 1 & 3 \\
\hline
\textbf{Total reviewed errors} & \textbf{164} & \textbf{212} \\
\hline
\end{tabular}
\caption{LeMAJ Error Analysis}
\label{tab:lemaj_errors}
\end{table*}

We provide in Table~\ref{tab:lemaj_errors} an error analysis for each of the steps in the pipeline:
\begin{itemize}
    \item On datapoints splitting only: In our first experiments, we iterated using reference data, i.e. a proprietary dataset which contained each answer split into individual data points by human legal experts. We then did some limited testing on how well the model was aligned with the human split and what the margins of error were.
    \begin{itemize}
        \item Our human reference data contained 2144 LDPs.
        \item LeMAJ (Claude Sonnet 3.5 v2) split the same dataset into 1964 LDPs, a difference of less than 10\%.
    \end{itemize}
    \item When looking at the end-to-end pipeline: When analysing the errors made by the LLM-as-a-Judge, we counted how many of these errors were due to different splitting by the human (reference data) and the LLM-as-a-Judge. Our error analysis showed that out of 212 tagging errors made by LeMAJ, 34 were due to difference in LDP split, or 16\% of all errors. We consider this an acceptable margin of error.
    \item We also performed a meta-review to account for the discrepancy between humans and the LLM. The meta-reviewers did not know which reviews came from LeMAJ and which reviews came from humans. Aside from that, they also had the ground truth at their disposal in case the reviews differed drastically.
\end{itemize}

\section{Appendix C. Datasets}
\label{appendix-datasets}

Below in Table \ref{tab:scoring_categories} is the grading mechanism used by lawyers to perform a fully 'manual' evaluation, i.e. without the involvement of LeMAJ or any LLM-as-a-Judge tool.
\begin{table*}[h]
    \centering
    \begin{tabular}{|l|c|l|c|}
        \hline
        \multicolumn{2}{|c|}{Correctness} & \multicolumn{2}{c|}{Relevance} \\
        \hline
        Category & Score & Category & Score \\
        \hline
        Completely correct & 1& Completely relevant & 1\\
        Mostly correct & 0.75& Mostly relevant & 0.75\\
        Equally correct and incorrect & 0.50& Equally relevant and irrelevant & 0.50\\
        Mostly incorrect & 0.25& Mostly irrelevant & 0.25\\
        Completely incorrect & 0& Completely irrelevant & 0\\
        \hline
    \end{tabular}
    \caption{Scoring Categories for Correctness and Relevance}
    \label{tab:scoring_categories}
\end{table*}

\section{Appendix D. LeMAJ Iterations}
\label{appendix:lemaj_iterations}

The below iterations have been performed solely on the proprietary dataset. The best model in terms of the trade-off between cost and performance was chosen to be represented in the final LegalBench evaluation. We present two accuracy metrics - a base and an adjusted version, with the adjusted version incorporating an additional text matching process alongside category evaluation.

\section{Appendix E. Baseline performance}

In our evaluation of base models without fine-tuning, Claude 3.5 Sonnet v2 demonstrated promising performance with an adjusted LeMAJ accuracy of 0.75. While the exact count of color tags did not precisely match the human evaluation, the overall proportion showed notable similarity, indicating a good baseline understanding of the task.

\begin{table*}[h!]
\centering
\begin{tabular}{|c|c|c|c|c}
\hline
EXP \# & Foundation Model & LeMAJ Accuracy & LeMAJ Accuracy Adjusted \\
\hline
Baseline 1-1 & Sonnet 3.5 v2 & 0.716 & 0.757 \\
Baseline 2-1 & Haiku & 0.469 & 0.447 \\
\hline
\end{tabular}
\caption{Baseline performance without finetuning}
\label{tab:baseline_no_ft}
\end{table*}

In contrast, Haiku scored 0.44 without fine-tuning, performing approximately 30\% worse than Sonnet. This performance gap is understandable given that Haiku is a significantly smaller model compared to Sonnet. However, a concerning pattern emerged in Haiku's output: the proportion of green tags was disproportionately high, and the model produced no red tags at all. This skewed distribution suggests that Haiku's base performance lacks the nuanced understanding required for accurate legal judgments. You can see those results in tables~\ref{tab:baseline_no_ft} and~\ref{tab:baseline_no_ft_ldp_split}.

\begin{table*}[h]
\centering
\small
\begin{tabular}{l p{2cm} c c c c c c}
\hline
\textbf{Experiment} & \textbf{Foundation Model} & \textbf{Correct} & \textbf{Incorrect} & \textbf{Irrelevant} & \textbf{Missing} & \textbf{Total Count} \\
\hline
Human & - & 901 & 24 & 362 & 857 & 2144 \\
Baseline 1-1 & Sonnet 3.5 v2 & 791 & 41 & 347 & 785 & 1964 \\
Baseline 1-2 & Haiku & 1229 & 0 & 177 & 445 & 1851 \\
\hline
\end{tabular}
\caption{Baseline LDP splitting without finetuning}
\label{tab:baseline_no_ft_ldp_split}
\end{table*}

Despite this, the results indicate substantial room for improvement through fine-tuning, particularly for the Haiku model, where targeted training could potentially address the imbalance in tag distribution and enhance overall performance.

\subsection{Finetuning hyperparameter search}

\textbf{Learning Rate Multiplier:} Learning Rate Multiplier (LRM) controls the maximum learning rate during the training process. When LRM is 1, the maximum learning rate during the training is 0.5. When LRM is 0.1, the maximum learning rate is 0.5 x 0.1 = 0.05. To be more specific, Claude 3 model customization on Amazon Bedrock employs a Piecewise Linear Learning Rate Scheduler, which starts with a learning rate of zero → gradually increases to a maximum learning rate during the first 5\% of training steps → remains constant until 80\% of steps are completed → proceeds to a linear cooldown to zero. Through extensive experimentation, we determined that a learning rate multiplier of 1.0 was optimal for our use cases.

\textbf{Epoch:} In contrast, the number of epochs demonstrated a more substantial influence on model performance. We observed a significant improvement in learning capabilities between epochs 2 and 3, with accuracy jumping from 0.694 in one experiment to 0.836 in another. Additionally, we noticed an increasing proportion of "Missing" tags compared to "Correct" tags as the number of epochs increased. However, it's crucial to emphasize that while epochs showed a more pronounced effect in our experiments, this is not universally applicable. The optimal number of epochs can vary depending on factors such as training set size, necessitating careful experimentation for each specific use case.

\subsection{Prompt iterations}

We explored a range of prompts, the results of which we can see in Table ~\ref{tab:prompt_iterations_ft}. Our findings revealed that prompt engineering can have a significant impact on the performance of baseline models. However, once the models were fine-tuned (EXP 1, 2 and 3), the differences in performance across various prompts became notably marginal.

\begin{table*}[h]
\centering
\small
\begin{tabular}{lccc p{2.8cm}}
\hline
\textbf{Experiment} & \textbf{Foundation Model} & \textbf{Prompt} & \textbf{LeMAJ Accuracy} & \textbf{LeMAJ Accuracy Adjusted} \\
\hline
Baseline 2-1 & Haiku & v1 & 0.469 & 0.447 \\
Baseline 2-2 & Haiku & v2 & 0.551 & 0.483 \\
Baseline 2-3 & Haiku & v3 & 0.564 & 0.515 \\
EXP1 & Haiku & v1 & 0.796 & 0.813 \\
EXP2 & Haiku & v2 & 0.805 & 0.821 \\
EXP3 & Haiku & v3 & 0.796 & 0.812 \\
\hline
\end{tabular}
\caption{Performance of prompt iterations}
\label{tab:prompt_iterations_ft}
\end{table*}

For instance, when applied to the baseline model, prompt v3 demonstrated the highest accuracy, followed by v2 and v1 respectively. Interestingly, this order shifted when the same prompts were applied to the fine-tuned model. In this scenario, prompt version 2 emerged as the top performer, achieving an accuracy of 0.821. These results suggest that while careful prompt design can enhance the performance of base models, the benefits of prompt engineering become less pronounced after fine-tuning, as the model adapts more comprehensively to the specific task at hand.

To better analyze what part of the information the judge would frequently misclassify, we look at the LDP tagging variability for each experiment in Table~\ref{tab:prompt_iterations_ft_ldp_split}. We see that some of the critical pieces it tends to misclassify are the "Incorrect" and "Missing" data points.

\begin{table*}[h]
\centering
\small
\begin{tabular}{l p{2cm} c c c c c p{1cm}}
\hline
\textbf{Experiment} & \textbf{Foundation Model} & \textbf{Prompt} & \textbf{Correct} & \textbf{Incorrect} & \textbf{Irrelevant} & \textbf{Missing} & \textbf{Total Count} \\
\hline
Human & - & - & 901 & 24 & 362 & 857 & 2144 \\
Baseline 2-1 & Haiku & v1 & 1229 & 0 & 177 & 445 & 1851 \\
Baseline 2-2 & Haiku & v2 & 1666 & 0 & 247 & 74 & 1987 \\
Baseline 2-3 & Haiku & v3 & 1250 & 3 & 190 & 425 & 1868 \\
EXP1 & Haiku & v1 & 1019 & 3 & 306 & 718 & 2046 \\
EXP2 & Haiku & v2 & 988 & 0 & 343 & 714 & 2045 \\
EXP3 & Haiku & v3 & 1018 & 1 & 326 & 651 & 1996 \\
\hline
\end{tabular}
\caption{LDP distribution for prompt iterations}
\label{tab:prompt_iterations_ft_ldp_split}
\end{table*}

\subsection{Data Augmentation}

Our hypothesis was that the misclassification was happening as a result of data skew in our training set and as a result we employed a set of data augmentation techniques meant to supplement some of the lacking LDPs in our training set (listed in Table~\ref{tab:data_augmentation_techniques}).

\begin{table*}[h!]
\centering
\begin{tabular}{p{0.3\textwidth}|p{0.6\textwidth}}
\hline
\textbf{Type} & \textbf{Description} \\
\hline
\texttt{remove\_info} & Find a "Correct" data point from the evaluation and remove the matching data point from the answer. Change its tag from "Correct" to "Missing". \\
\hline
\texttt{incomplete\_info} & Find a "Correct" data point from the evaluation and modify the matching data point from the answer so that the information it conveys becomes incomplete. Change its tag from "Correct and Relevant" to "Missing". \\
\hline
\texttt{change\_value} & Modify a specific number or named entity in the answer and tag it with "Incorrect" in the evaluation.\\
\hline
\texttt{add\_extra\_info} & Add 1-2 sentences to the answer using LLM's own legal knowledge and tag it with "Incorrect" in the evaluation. \\
\hline
\texttt{contradicting\_info} & Rewrite the answer so that it contradicts with the ground truth. Keep the original data points in the evaluation as "Missing" and add the rewritten data point with "Incorrect" tags. \\
\hline
\end{tabular}
\caption{Data augmentation process}
\label{tab:data_augmentation_techniques}
\end{table*}

Contrary to our expectations, models fine-tuned with the augmented dataset showed a slight decrease in accuracy (Table~\ref{tab:data_augmentation_ft}). For instance, the non-augmented model (EXP2-1) achieved a score of 0.821, while the "Incorrect, Missing" augmented model (EXP9) scored 0.798. However, the augmented models did return more "Incorrect" and "Missing" samples (Table~\ref{tab:data_augmentation_ft_ldp}), suggesting that they were learning the distribution present in the training set.

\begin{table*}[h!]
\centering
\small
\begin{tabular}{|c|c|c|c|>{\centering\arraybackslash}p{2.8cm}|}
\hline
EXP \# & Foundation Model & Augmentation & LeMAJ Accuracy & LeMAJ Accuracy Adjusted \\
\hline
EXP1 & Haiku & None & 0.80513 & 0.82148 \\
EXP2 & Haiku & Incorrect & 0.77439 & 0.80571 \\
EXP3 & Haiku & Incorrect, Missing & 0.78164 & 0.79887 \\
EXP4 & Haiku & None & 0.80695 & 0.83697 \\
EXP5 & Haiku & Incorrect (n=10) & 0.79519 & 0.81477 \\
EXP6 & Haiku & Incorrect (n=15) & 0.81386 & 0.82679 \\
\hline
\end{tabular}
\caption{Performance with different augmentations}
\label{tab:data_augmentation_ft}
\end{table*}

\begin{table*}[h!]
\centering
\small
\begin{tabular}{|c|p{1.5cm}|p{1.5cm}|c|c|c|c|>{\centering\arraybackslash}p{1.5cm}|}
\hline
EXP \# & Foundation Model & Augmentation & Correct & Incorrect & Irrelevant & Missing & Total Count \\
\hline
Human & - & - & 901 & 24 & 362 & 857 & 2144 \\
EXP1 & Haiku & None & 988 & 0 & 343 & 714 & 2045 \\
EXP2 & Haiku & Incorrect & 1046 & 23 & 338 & 669 & 2076 \\
EXP3 & Haiku & Incorrect, Missing & 1006 & 7 & 350 & 720 & 2083 \\
EXP4 & Haiku & None & 841 & 1 & 433 & 888 & 2163 \\
EXP5 & Haiku & Incorrect (n=10) & 869 & 4 & 372 & 802 & 2047 \\
EXP6 & Haiku & Incorrect (n=15) & 821 & 12 & 376 & 917 & 2126 \\
\hline
\end{tabular}
\caption{LDP distribution with different augmentations}
\label{tab:data_augmentation_ft_ldp}
\end{table*}

These data augmentation experiments suggest that while data augmentation is a popular technique in LLM trainings generally, its application in the legal domain may be challenging and require extensive human evaluation. However, from the customer's perspective, it would be more sensible to focus on annotating the available training set rather than reviewing synthetic samples. Thus, data augmentation should be considered as a last resort when no other data is available.

\subsection{LLM Jury}

In our pursuit of optimizing the LLM Judge model's performance, we explored several ensemble approaches, collectively referred to as the LLM Jury. These methods aim to leverage the strengths of multiple fine-tuned models to enhance overall accuracy and robustness. We tried four different flavors of LLM Jury:

\textbf{Rule-based:} This approach employs heuristics to determine the final verdict. This method assigns greater weight to non-green labels, prioritizing them in the order of red, grey, orange, and green, in accordance with the customer's emphasis on detecting incorrect (red) data points.

\textbf{Majority Voting:} This approach compares the outputs of three judges and selects the most common color label for each data point.

\textbf{Rule-based + Majority Voting:} Building upon the aboves, we developed a hybrid approach that combines majority voting with rule-based decision-making. This method prioritizes red labels when identified by any judge, otherwise defaulting to the majority rule.

\textbf{Chain-of-verification:} This approach, utilizing a base Claude 3.5 Sonnet v2 as a final judge to refine the output generated by the fine-tuned model. This method leverages the LLM's inherent reasoning and self-verification capabilities.

\begin{table*}[h!]
\centering
\small
\begin{tabular}{p{2.2cm}cccccc}
\hline
\textbf{Jury Method} & \textbf{Accuracy} & \textbf{Correct} & \textbf{Incorrect} & \textbf{Irrelevant} & \textbf{Missing} & \textbf{Total Count} \\
\hline
Human & - & 901 & 24 & 362 & 857 & 2144 \\
Rule Based & 0.842 & 841 & 1 & 452 & 916 & 2210 \\
Majority Voting & 0.854 & 978 & 1 & 446 & 793 & 2218 \\
Majority Voting + Rule Based & 0.852 & 912 & 3 & 447 & 847 & 2209 \\
Majority Voting + Rule Based & 0.845 & 858 & 20 & 444 & 924 & 2246 \\
Chain of Verification & 0.806 & 830 & 1 & 423 & 841 & 2095 \\
\hline
\end{tabular}
\caption{LLM Jury accuracy and LDP splits}
\label{tab:llm_jury_ft}
\end{table*}

Among these four approaches, the combination of majority voting and rule-based decision-making demonstrated the best overall performance (0.852). Notably, we discovered that by combining two models with "Incorrect" data augmentation and one without, we could maximize the ability to discern "Incorrect" cases. This configuration, corresponding to the fifth row in Table~\ref{tab:llm_jury_ft}, achieved an accuracy of 0.844, only 0.1\% lower than the best-performing method while generating significantly more red samples.

\section{Appendix F. Examples of Legal Data Points}
\label{appendix-exemples-of-legal-data-point}

\begin{figure*}
    \centering
    \includegraphics[width=1.1\linewidth]{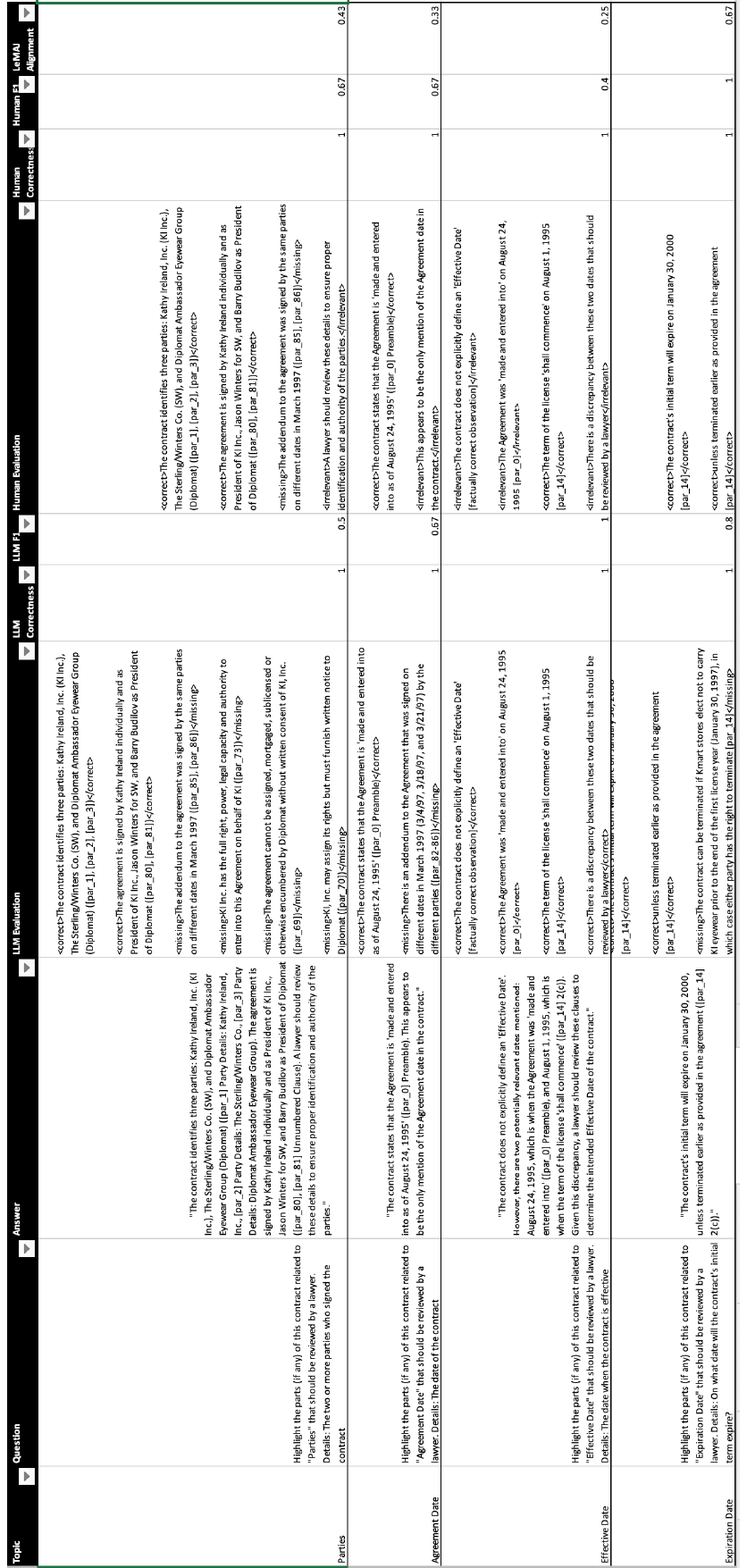}
    \caption{An example of LDPs with both the LLM evaluation performed by LeMAJ and the human evaluation by a human legal expert, resulting in the LeMAJ Alignment score.}
    \label{fig:enter-label}
\end{figure*}

We provide in Table~\ref{tab:ldp-examples} and Figure \ref{fig:enter-label} several examples of pairs of questions and answers, as well as the evaluation of each answer using the LeMAJ framework: spitting an answer into LDPs, and evaluating each one of them.
\begin{table*}
\centering
\small
\begin{tabular}{|p{2cm}|p{3cm}|p{3.5cm}|p{3.5cm}|}  
\hline
\textbf{Issue} & \textbf{Question} & \textbf{Answer} & \textbf{Evaluation}\\
\hline
Document Name & Highlight the parts (if any) of this contract related to "Document Name" that should be reviewed by a lawyer. Details: The name of the contract &  The name of the contract is 'WEB SITE HOSTING AGREEMENT' ([par\_1] Preamble).  & \textless \textbf{correct}\textgreater The name of the contract is "WEB SITE HOSTING AGREEMENT" [par\_1] \\
\hline
Agreement Date & Highlight the parts (if any) of this contract related to "Agreement Date" that should be reviewed by a lawyer. Details: The date of the contract &  The agreement date is April 6, 1999 ([par\_1] Unnumbered Clause).  & \textless \textbf{correct}\textgreater The agreement date is April 6, 1999 ([par\_1])

\textless \textbf{missing}\textgreater The agreement also mentions April 1, 1999 as the commencement date ([par\_36]) \\
\hline
Effective Date & Highlight the parts (if any) of this contract related to "Effective Date" that should be reviewed by a lawyer. Details: The date when the contract is effective & The contract specifies that it is entered into on April 6, 1999 ([par\_1] 1). This date appears to be the effective date of the contract, though the term 'Effective Date' is not explicitly defined or used in the agreement. & \textless \textbf{correct}\textgreater The contract is entered into on April 6, 1999 [par\_1]

\textless \textbf{irrelevant}\textgreater This date appears to be the effective date of the contract, though the term 'Effective Date' is not explicitly defined or used in the agreement.\\
\hline
Governing Law & Highlight the parts (if any) of this contract related to "Governing Law" that should be reviewed by a lawyer. Details: Which state/country's law governs the interpretation of the contract? &  The contract specifies that the laws of the State of Florida govern its validity, construction, interpretation, and legal effect ([par\_46] 46).& \textless \textbf{correct}\textgreater The contract is governed by the laws of the State of Florida [par\_46]

\\
\hline
\end{tabular}
\caption{Examples of Q\&A pairs with an evaluation of each answer following the LeMAJ framework. These evaluations can be done automatically or by a domain expert.}
\label{tab:ldp-examples}
\end{table*}

\section{Appendix G. DeepEval Correctness Prompt}
\label{appendix-deepeval-prompt}

We use the following prompt with the G-Eval method of DeepEval to assess the correctness of an answer:
\textit{
You are a legal expert, tasked with evaluating an answer to a question about a legal contract.
You have been provided with the following information:
\begin{itemize}
    \item  A legal contract
    \item A question about the legal contract
    \item An answer to the question
\end{itemize}
Evaluate the correctness of the answer.}

\section{Appendix H. Commercial use case: reduce human review efforts through triage}
\label{appendix-commercia-application}

\subsection{Time savings on our proprietary dataset}

We applied the thresholds outlined above to obtain the breakdown in Table \ref{table:qa-review}. For verification purposes, the table also includes the Correctness and Relevance scores as verified by human review. We can see that a sizable chunk of answers pass both thresholds, while their average Correctness and Relevance scores remain very high. In addition, the LeMAJ Alignment score gives us confidence that LeMAJ has evaluated these answers in a manner very closely aligned to how humans would have. We can therefore assume that this portion of answers have been evaluated correctly and do not need human review.

\begin{table*}[h]
    \centering
    \small
    \begin{tabular}{l*{9}{c}}
        \hline
        & \rotatebox{50}{Leases} & \rotatebox{50}{LPA} & \rotatebox{50}{MSAs} & \rotatebox{50}{NDA} & \rotatebox{50}{SaaS} & \rotatebox{50}{SHAs} & \rotatebox{50}{Side Letters} & \rotatebox{50}{SPAs} & \rotatebox{50}{Supplier} \\
        \hline
        QA pairs & 100 & 120 & 100 & 100 & 95 & 80 & 90 & 85 & 100 \\
        Passing triage & \textbf{30} & \textbf{15} & \textbf{30} & \textbf{46} & \textbf{22} & \textbf{25} & \textbf{36} & \textbf{32} & \textbf{34} \\
        Human Correctness & 1 & 1 & 1 & 0.99 & 0.97 & 1 & 1 & 0.99 & 0.99 \\
        Human Relevance & 0.91 & 0.95 & 0.95 & 0.91 & 0.92 & 0.95 & 0.97 & 0.96 & 0.93 \\
        LeMAJ Alignment & 0.86 & 0.91 & 0.81 & 0.84 & 0.83 & 0.93 & 0.86 & 0.94 & 0.86 \\
        QA pairs to review & \textbf{70} & \textbf{105} & \textbf{70} & \textbf{54} & \textbf{73} & \textbf{55} & \textbf{54} & \textbf{53} & \textbf{66} \\
        \hline
    \end{tabular}
    \caption{Triaging of answers with Correctness = 1 and Relevance $\geq$ 0.80}
    \label{table:qa-review}
\end{table*}

To translate that into time savings, we measured time spent on reviewing a set of answers by contract type, as summarized in the Table \ref{table:qa-human-time}, and calculated how the triaging would affect time spent on reviewing the remainder of answers.

\begin{table*}[h]
    \centering
    \small
    \begin{tabular}{p{3cm}|*{9}{c}}
        \hline
        & \rotatebox{50}{Leases} & \rotatebox{50}{LPA} & \rotatebox{50}{MSAs} & \rotatebox{50}{NDA} & \rotatebox{50}{SaaS} & \rotatebox{50}{SHAs} & \rotatebox{50}{Side Letters} & \rotatebox{50}{SPAs} & \rotatebox{50}{Supplier} \\
        \hline
        \textbf{QA pairs} & 100 & 120 & 100 & 100 & 100 & 80 & 90 & 85 & 100 \\
        \hline
        \textbf{Approx. human review time (hours)}& 7.5 & 15 & 10 & 10 & 10 & 14.5 & 15 & 15 & 10 \\
        \hline
        \textbf{Proportion to review} & 30\% & 14\% & 30\% & 50\% & 20\% & 40\% & 40\% & 40\% & 30\% \\
        \hline
        \textbf{New estimated human review time (hours)} & $\sim$5 & $\sim$13 & $\sim$7 & $\sim$5 & $\sim$8 & $\sim$11 & $\sim$9 & $\sim$9 & $\sim$7 \\
        \hline
        
    \end{tabular}
    \caption{QA Pairs and Human Review Time Distribution}
    \label{table:qa-human-time}
\end{table*}

\subsection{Time savings on LegalBench}

On average, a manual review by a human legal expert of the LegalBench dataset would take a little under eight hours, with around 3 to 4 minutes on average spent per question (this is of course an average and can vary significantly depending on the complexity of the question). We applied thresholds of a Correctness score of 1 \textit{\textbf{and}} a Relevance score of at least 0.85, resulting in the triage included in Table \ref{tab:evaluation_qa_pairs} below. Rather than include a breakdown per contract type, we have included the same review, performed by different reviewers, to illustrate consistency in the results of our approach.

\begin{table*}[t]
    \centering
    \small
    \begin{tabular}{|l|c|c|}
        \hline
        \textbf{Evaluation by} & \textbf{Reviewer A} & \textbf{Reviewer B} \\
        \hline
        QA pairs& 150 & 150 \\
        \hline
        QA pairs passing triage& 51& 51\\
        \hline
        LeMAJ Alignment of triaged QA pairs& 0.72 & 0.78 \\
        \hline
        QA pairs to review& 99 & 99 \\
        \hline
        LeMAJ Alignment of QA pairs to review& 0.49 & 0.5 \\
        \hline
    \end{tabular}
    \caption{Triaging of answers based on Correctness and Relevance thresholds on LegalBench }
    \label{tab:evaluation_qa_pairs}
\end{table*}

\begin{table*}[t]
    \centering
    \small
    \begin{tabular}{|l|c|c|}
    \hline
    \textbf{Evaluation by} & \textbf{Reviewer A} & \textbf{Reviewer B} \\ \hline
    QA pairs & 150 & 150 \\ \hline
    Time spent (in hours) & 8.25 & 7.55 \\ \hline
    Answers to review & 99 & 99 \\ \hline
    New estimated human review time & 5.45 & 4.98 \\ \hline
    Approx. time saving & 30\%& 30\% \\ \hline
    \end{tabular}
    \caption{Illustrative potential time savings on LegalBench}
    \label{tab:evaluation_metrics}
\end{table*}

A human review of those triaged answers reveals an average Correctness score of 0.96 and an average Relevance score of 0.86, remaining very close to LeMAJ scores. 
We are then able to create an approximation of time savings that this triaging would allow. In a commercial context, such as that of an in-house legal team or a legal AI team, this methodology can save reviewing teams time and effort spent on exercises like quality assessments of answers and legal review.

\end{document}